\documentclass[10pt,twocolumn,letterpaper]{article}

\usepackage[pagenumbers]{cvpr} %

\usepackage[HTML]{xcolor}

\newcommand{\descrot}{\colorbox[HTML]{c999cc}{RotDesc\&Match}}
\newcommand{\rotmatch}{\colorbox[HTML]{f06292}{RotMatch}}
\newcommand{\norot}{\colorbox[HTML]{7bc8a4}{NoRot}}

\usepackage{amsmath}
\usepackage{amssymb}
\usepackage{amsfonts}
\usepackage{mathabx}

\definecolor{cvprblue}{rgb}{0.21,0.49,0.74}
\usepackage[pagebackref,breaklinks,colorlinks,citecolor=cvprblue]{hyperref}

\usepackage{colortbl}
\usepackage{pifont}

\usepackage{multirow}
\usepackage{bigdelim}

\def\paperID{{12}} %
\def\confName{IMW:CVPR}
\def\confYear{2026}

\title{Who Handles Orientation? Investigating Invariance in Feature Matching}

\author{
David Nordström$^{1}$   
\and 
Johan Edstedt$^{2}$
\and
Fredrik Kahl$^{1}$    
\and
Georg Bökman$^{3}$%
\and 
\\
$^1$Chalmers University of Technology \hspace{2em} 
$^2$Linköping University
\hspace{2em} 
$^3$University of Amsterdam
}

\begin{document}
\maketitle
\begin{abstract}
Finding matching keypoints between images is a core problem in 3D computer vision. However, modern matchers struggle with large in-plane rotations. A straightforward mitigation is to learn rotation invariance via data augmentation. However, it remains unclear at which stage rotation invariance should be incorporated. In this paper, we study this in the context of a modern sparse matching pipeline. We perform extensive experiments by training on a large collection of 3D vision datasets and evaluating on popular image matching benchmarks. Surprisingly, we find that incorporating rotation invariance already in the descriptor yields similar performance to handling it in the matcher. However, rotation invariance is achieved earlier in the matcher when it is learned in the descriptor, allowing for a faster rotation-invariant matcher.
Further, we find that enforcing rotation invariance does not hurt upright performance when trained at scale.
Finally, we study the emergence of rotation invariance through scale and find that increasing the training data size substantially improves generalization to rotated images.
We release two matchers robust to in-plane rotations that achieve state-of-the-art performance on e.g.\ multi-modal (WxBS), extreme (HardMatch), and satellite image matching (SatAst). Code is available \href{https://github.com/davnords/loma}{here}.

\end{abstract}

\section{Introduction} \label{sec:intro}

\begin{figure*}[t]
  \centering
  \begin{subfigure}{0.49\textwidth}
    \centering
    \includegraphics[width=\linewidth]{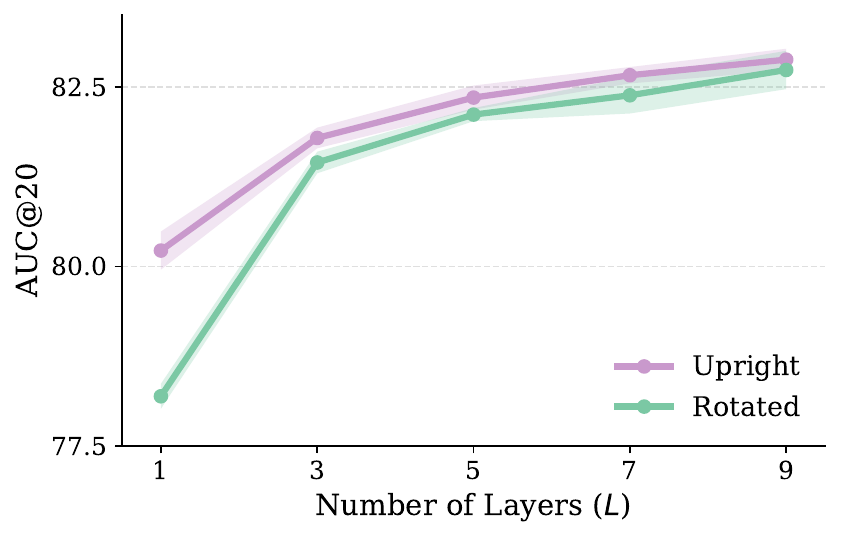}
    \caption{Rotation invariance in only the matcher (\textit{\rotmatch}).}
  \end{subfigure}
  \hfill
  \begin{subfigure}{0.49\textwidth}
    \centering
    \includegraphics[width=\linewidth]{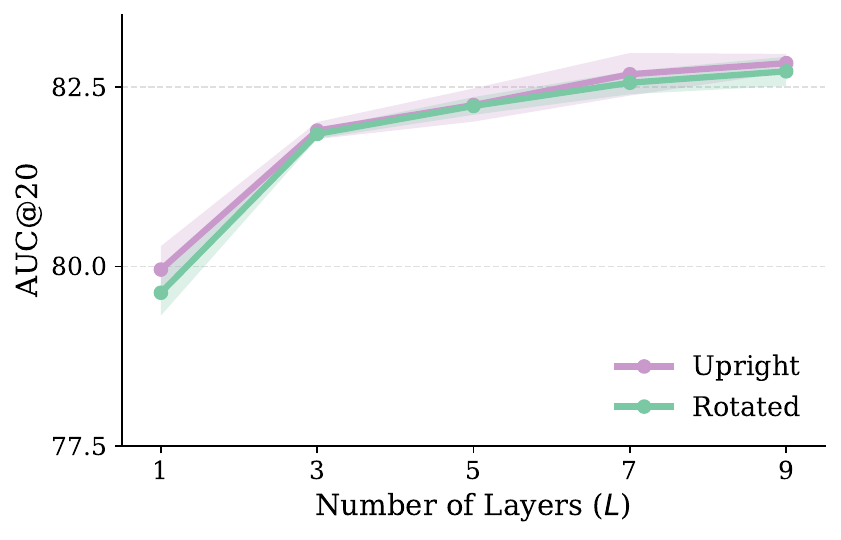}
    \caption{Rotation invariance in both descriptor and matcher (\textit{\descrot}).}
  \end{subfigure}
  \caption{\textbf{Who Handles the Orientation?} We compare incorporating rotation invariance in (a) the matcher and (b) both the matcher and descriptor, by reporting the AUC@20 on MegaDepth-1500~\cite{li2018megadepth, sun2021loftr} (rotated and upright) for different stopping layers using $N=2048$ keypoints. We average over 5 runs and include a 95\% confidence interval.}
  \label{fig:teaser}
\end{figure*}

A crucial part of both visual localization and Structure-from-Motion (SfM)~\cite{hartley2003multiple} is finding correspondences between a query image and another image or a map. This is typically achieved in a three-step manner: (i) detection of sparse keypoints, (ii) description of their local appearance (e.g., SIFT~\cite{lowe2004distinctive}), and (iii) matching (e.g., mutual nearest neighbor). Learning-based approaches have been proposed for each stage of this pipeline. Sparse matchers, such as SuperGlue~\cite{sarlin2020superglue} and LightGlue~\cite{lindenberger2023lightglue}, replace nearest neighbor search with a graph neural network in order to match descriptors. Recently, \citet{loma:2026} showed that scaling data and compute yields substantial performance gains for sparse matchers. The resulting model, LoMa, achieves state-of-the-art (SotA) performance, even surpassing dense matchers on some benchmarks.

Many downstream tasks require robustness to in-plane rotations, including aerial imagery, handheld mapping, and medical imaging. These scenarios often lack a canonical orientation. For instance, several categories in the recently introduced HardMatch benchmark~\cite{loma:2026}, such as star constellations (\cf \cref{fig:hardmatch-qual}), appear in arbitrary rotations. Yet, SotA sparse matchers~\cite{sarlin2020superglue,lindenberger2023lightglue,loma:2026} do not incorporate rotational symmetry into their training. While this follows common practice in modern vision pipelines, \eg~\cite{deitiii:2022, wang2025vggt,siméoni2025dinov3}, it leads to performance degradation on several HardMatch categories (\cf \cref{fig:radar}), satellite image matching (\cf~\cref{tab:challenging-matching}), and rotated versions of popular matching benchmarks (\cf~\cref{tab:relpose}).

Rotation invariance in image matching is a popular research topic, especially at the detector and descriptor stages~\cite{Zhao2023ALIKED, edstedt2025dad, lee2022self, lee2023learning, bökman2024steerers}, however one understudied aspect is the rotation invariance of sparse matchers.
In particular, the question of which part of the pipeline should handle the orientation remains open.

We provide an empirical study on the effect of rotation invariance at different stages of a modern sparse matching pipeline. We do so by training three different models: (i)~without rotation augmentation, (ii)~with rotation augmentation when training the matcher, and (iii)~with rotation augmentation for both the matcher and descriptor. Through extensive experimentation, we address the following research questions: 

\begin{enumerate}%
\item \textbf{At which stage should invariance be incorporated?} We investigate whether rotation invariance should be included already in the descriptor or delegated to the matcher. While we observe no significant difference in later layers (\cf \cref{tab:relpose,tab:challenging-matching,tab:hardmatch}), we find that including rotation invariance in the descriptor
allows earlier stopping of the matcher, enabling faster rotation-invariant matching (\cf \cref{fig:teaser}). This supports incorporating invariance already at the descriptor stage. 
\item \textbf{Does invariance hurt performance on non-rotated benchmarks?} We observe a slight deterioration in performance on standard benchmarks (\cf \cref{tab:relpose}). However, generalization to out-of-distribution upright matching tasks marginally improves (\cf \cref{tab:challenging-matching,tab:hardmatch}), leading to SotA upright performance on \eg WxBS. This suggests that rotation invariance enhances robustness. 
\item \textbf{Does robustness to rotations emerge through data?} 
By comparing models trained only on MegaDepth with those trained on our full dataset, we find the latter to be significantly more robust to in-plane rotations (\cf \cref{tab:more-data}), even without rotation augmentation. Notably, AUC@20 improves by 52 points on the rotated evaluation set of MegaDepth-1500.
Hence, robustness to in-plane rotations can emerge naturally when training on diverse data, without explicit rotation augmentation.
\end{enumerate}

\section{Related Work} \label{sec:related}

Classically, pixel correspondences between two images can be found by detecting salient keypoints, describing them locally as high-dimensional feature vectors, and matching them using mutual nearest neighbor search. Detection and description have traditionally relied on hand-crafted methods. The features are described in a local rotation frame, making them rotation-invariant, \eg~SIFT~\cite{lowe2004distinctive}, SURF~\cite{bay2006surf}, and ORB~\cite{rublee2011orb}. Subsequently, learning-based approaches gained popularity for detection and description. Notable early works include TILDE~\cite{verdie2015tilde}, SuperPoint~\cite{detone2018superpoint}, and AffNet~\cite{mishkin2018repeatability}. Subsequently, graph neural network approaches (\eg~SuperGlue~\cite{sarlin2020superglue} and LightGlue~\cite{lindenberger2023lightglue}) were proposed to improve descriptor matching. Recently, LoMa~\cite{loma:2026} showed that sparse matchers could rival their dense counterparts~\cite{edstedt2024roma, zhang2025ufm, edstedt2025romav2harderbetter} through an improved pipeline with increased data and compute. However, a major weakness of these approaches is the lack of robustness to camera rotation.

One way to obtain rotation invariance is by using equivariant neural networks~\cite{cohen2016group,weiler2023EquivariantAndCoordinateIndependentCNNs}. They have been included in the detector~\cite{lee2021self,lee2022self} and for joint detection and description~\cite{yi2016lift,lee2023learning}, as well as in the feature extraction part of semi-dense matchers~\cite{bokman2022case}.
Another straightforward approach is to encourage the networks to learn symmetries rather than hard-coding them. Recently, Steerers~\cite{bökman2024steerers,bokman2024affine} proposed to use a latent transformation (steerer) for aligning descriptions of corresponding keypoints in rotated images.
The simplest steerer would be the identity transform, and this corresponds to invariant descriptors trained via data augmentation, which has been explored in \eg~ALIKED~\cite{Zhao2023ALIKED}.

However, in SotA sparse matchers~\cite{sarlin2020superglue, lindenberger2023lightglue,loma:2026} rotation invariance via any of the above methods remains unstudied.

\section{Method}

\begin{figure*}[t]
    \centering
        \includegraphics[width=0.95\linewidth]{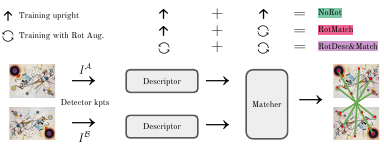}

    \caption{\textbf{Experimental setup.} We train descriptors and matchers with and without rotation augmentation, resulting in the three different models used for our main experiments. NoRot corresponds to the LoMa-B baseline.}
    \label{fig:method-figure}
\end{figure*}

In this section, we begin by providing preliminaries on matching (\cref{subsec:problem}), the rotation augmentation scheme (\cref{subsec:rotaug}), our models (\cref{subsec:models}), and training data (\cref{subsec:training-data}). Lastly, we outline the training setup (\cref{subsec:training}). We provide a schematic view of our experimental setup in \cref{fig:method-figure}. Our work follows~\cite{loma:2026}, with the addition of rotation augmentation at different stages of the pipeline. We will make the training and evaluation code publicly available.

\subsection{Preliminaries on Matching}\label{subsec:problem}

The task of image matching is to find \textit{matching} keypoints between two images $I^{\mathcal{A}}$ and $I^\mathcal{B}$. The term \textit{matching} is defined as pixel correspondences relating to the same 3D point. Broadly, there are two paradigms: (i) sparse and (ii) dense, detector-free matching. The latter uses no explicit keypoints and instead finds dense or semi-dense features across the image. This includes matchers such as LoFTR~\cite{sun2021loftr}, RoMa~\cite{edstedt2024roma}, and MASt3R~\cite{duisterhof2025mast3r}. Detector-free matching is computationally heavy as it is done per-pixel. In this work, we instead consider sparse matching.

Typically, sparse image matching follows a three-step process. Keypoints $x_i^{\mathcal{A}}$ and $x_j^{\mathcal{B}}$ are first detected independently in each image, where $i=1,\ldots,N$ and $j=1,\ldots,M$, then described by local features $d_i^\mathcal{A}$ and $d_j^\mathcal{B}$, and finally keypoint correspondences are obtained by matching the descriptors. Traditionally, keypoint descriptors $d_i^\mathcal{A}$ and $d_j^\mathcal{B}$ were obtained using methods such as SIFT and matched via mutual nearest neighbor search. Modern 3D computer vision has embraced learning-based methods for this three-step process. The keypoints are detected by the \textit{detector} $f_\omega$, described by the \textit{descriptor} $g_\theta$, and matched by the \textit{matcher} $h_\phi$, where $\{\omega, \theta, \phi\}$ are learnable parameters from neural networks. 

The matcher $h_\phi$ takes keypoints and descriptors from both images as input and produces refined descriptors $\tilde{d}$ through $L$ layers. The resulting descriptors, which now depend on both images, are used to compute the similarity matrix $S_{ij} = (\tilde{d}_i^{\mathcal A})^\top \tilde{d}_j^{\mathcal B}$. Soft assignments $\mathcal{P}$ are produced by taking the double-softmax of $S$. A correspondence $(i,j)$ is established when $\mathcal{P}_{ij}$ is the maximum in both its row and column, \ie when the match is mutual. Matches with $\mathcal{P}_{ij} < \tau$ are discarded.

\subsection{Rotation Augmentation}\label{subsec:rotaug}
To investigate the effect of rotation invariance, we include random rotation augmentations in some experiments. To do so, we apply independent in-plane rotations to each image in the pair. Specifically, we sample rotation angles $\alpha^{\mathcal{A}}, \alpha^{\mathcal{B}} \in \{0^\circ, 90^\circ, 180^\circ, 270^\circ\}$ and rotate the images and their associated depth maps accordingly. Rotations $\alpha^\mathcal{A}$ and $\alpha^\mathcal{B}$ are sampled independently. The camera intrinsics are updated to remain consistent with the rotated image coordinates. In \cref{subsec:ablations}, we study the effect of applying joint rotations (\ie $\alpha^\mathcal{A}=\alpha^\mathcal{B}$) during training of the descriptor.

\subsection{Models}\label{subsec:models}

Our setup follows LoMa~\cite{loma:2026}. That is, we first train the descriptor $g_\theta$ and then freeze $g_\theta$ to supervise the matcher $h_\phi$. We do not train the detector $f_\omega$. Instead, we use DaD~\cite{edstedt2025dad} to supervise $g_\theta$ and $h_\phi$ by sampling $N=M=2048$ keypoints. Both networks are trained on a large data mix (see \cref{subsec:training-data}). We use DeDoDe-G~\cite{edstedt2024dedode, edstedt2024dedodev2} as $g_\theta$ and LightGlue as $h_\phi$. This is equivalent to the architecture of LoMa-B in~\cite{loma:2026}. Differing only by the training procedure, we consider three models (schematically visualized in \cref{fig:method-figure}):
\begin{enumerate}
    \item \textbf{\norot:} Both descriptor and matcher trained on upright images. The performance is similar to LoMa-B~\cite{loma:2026}.
    \item \textbf{\rotmatch:} The matcher is trained with rotation augmentation while using the same descriptor as NoRot.
    \item \textbf{\descrot:} Both the descriptor and matcher are trained with rotation augmentation.
\end{enumerate}

Training only the descriptor with rotation augmentation does not yield a rotation-invariant matching pipeline as the LightGlue architecture includes the image coordinates as input to the sparse matcher.
Therefore, we do not include this case in our study.

\begin{figure}[t]
    \centering
        \includegraphics[width=0.95\linewidth]{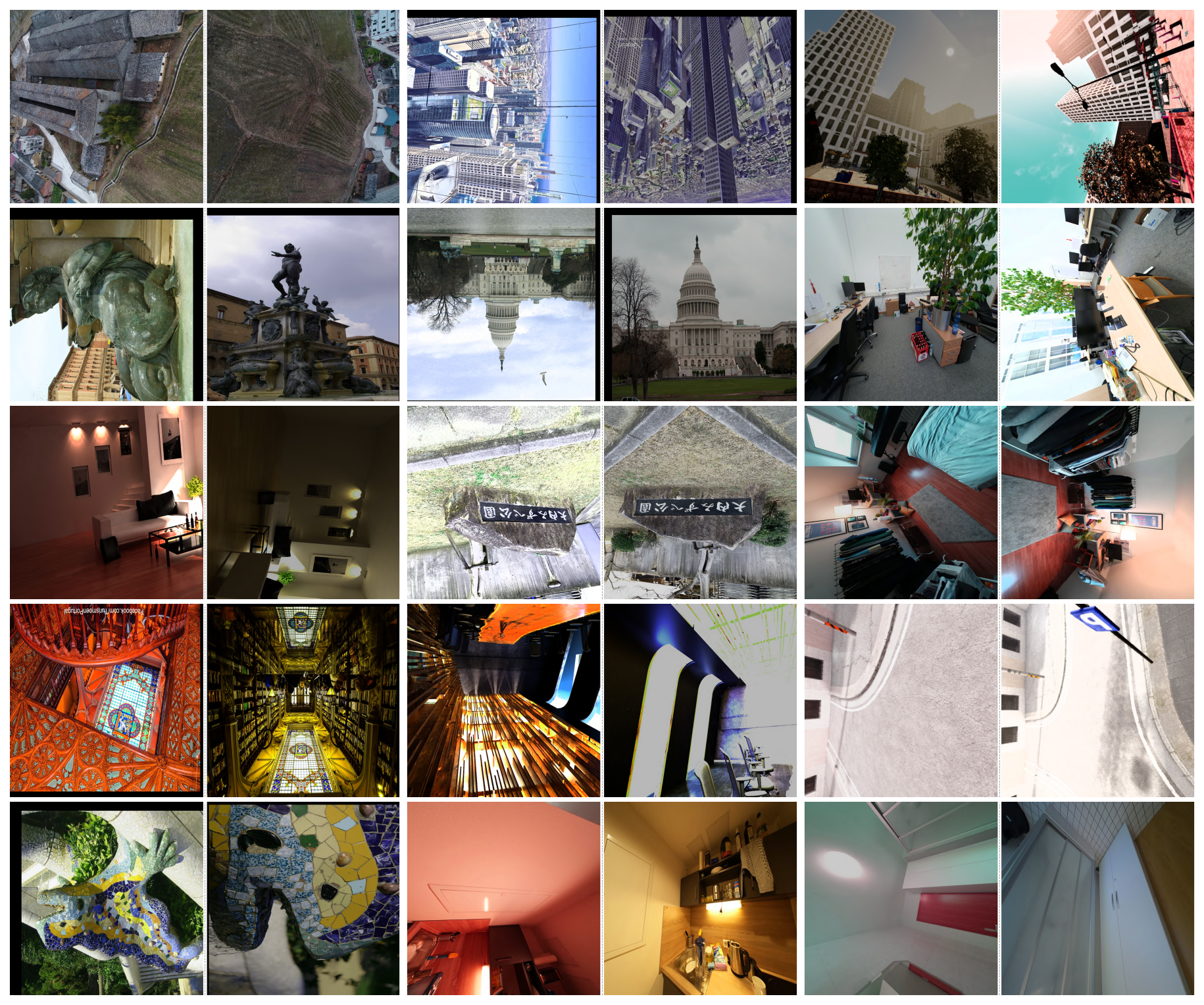}

    \caption{\textbf{Training batch.} Illustration of a training batch with rotation augmentation.}
    \label{fig:training-batch}
\end{figure}

\subsection{Training Data}\label{subsec:training-data}

We use the same collection of seventeen 3D datasets as in \cite{loma:2026} for all experiments, if not otherwise stated. The data mix, presented in \cref{tab:dataset_mix}, includes both real data (from \eg~multi-view stereo pipelines) and synthetic sources. In \cref{fig:training-batch}, we illustrate a random training batch with rotation augmentation applied, as is done in the training of our rotation-invariant models.

\begin{table}[ht] \centering
\small
\caption{\textbf{Training data}. For the main experiments we use a large collection of 3D datasets for training.}
\label{tab:dataset_mix}
\setlength{\tabcolsep}{2pt}
\begin{tabular}{lcc}
\hline
\toprule
Datasets                          & Type / GT Source            & Weight \\ 
\midrule
ScanNet++ v2~\cite{yeshwanth2023scannet++} & Indoor / Mesh & 1 \\%\hdashline
BlendedMVS~\cite{yao2020blendedmvs}  & Aerial / Mesh      &1 \\
Map-Free~\cite{arnold2022map}& Object-centric / MVS &1 \\
Hypersim~\cite{roberts2021hypersim}& Indoor / Graphics          &1 \\
MegaScenes~\cite{tung2024megascenes} & Outdoor / MVS & 1 \\
MegaDepth~\cite{li2018megadepth}    & Outdoor / MVS        &1 \\
MegaDepth (Re-MVS) & Outdoor / MVS        &1 \\
AerialMD~\cite{vuong2025aerialmegadepth}    & Aerial / MVS        &1 \\
TartanAir v2~\cite{wang2020tartanair}& Outdoor / Graphics          &1 \\

MPSD~\cite{mpsd:2020} & Driving / MVS  & 0.1 \\
ASE~\cite{AriaSynthEnv:2025} & Indoor / Graphics & 0.1\\
CO3Dv2~\cite{co3d:2021} & Object-centric / MVS & 0.1 \\
MegaSynth~\cite{Jiang_2025_CVPR} & Indoor / Graphics & 0.1\\
SpatialVID~\cite{wang2025spatialvid} & Forward-motion / MVS & 0.01\\
\midrule
FlyingThings3D~\cite{mayer2016large} & Outdoor / Graphics & 0.5\phantom{0}\\
UnrealStereo4k~\cite{tosi2021unrealstereo4k} & Outdoor / Graphics & 0.01\\
Virtual KITTI 2~\cite{gaidon2016virtual,cabon2020vkitti2} & Outdoor / Graphics & 0.01\\

\bottomrule
\end{tabular}
\vspace{-0.5em}
\normalsize
\end{table}

\subsection{Training}\label{subsec:training}

We follow the LoMa training protocol~\cite{loma:2026}. That is, we use a global batch size of 64, a learning rate of $2\times10^{-4}$, and the AdamW~\cite{loshchilov2018decoupled} optimizer. The descriptor and matcher are trained for 50k and 100k steps, respectively. All experiments are run on a single compute node with $8$ A100 GPUs. Training takes approximately a day.

\section{Experiments}

In this section, we compare our three model variants on a wide range of benchmarks using both upright and rotated image pairs. We start with relative pose estimation (\cref{subsec:relpose}), followed by a collection of image matching evaluations (\cref{subsec:wxbs,subsec:satast,subsec:hardmatch}). Lastly, we perform ablation studies (\cref{subsec:ablations}) and analyze qualitative differences in the descriptors (\cref{subsec:desc-qual}). We further include comparisons to SuperPoint+SuperGlue (SP+SG), SuperPoint+LightGlue (SP+LG), and ALIKED+LightGlue (ALIKED+LG) where appropriate. Following \cite{loma:2026}, all evaluations use $N=4096$ maximum keypoints. Except for the ablations, we use random independent rotations $\alpha^\mathcal{A}, \alpha^\mathcal{B} \in \{0^\circ, 90^\circ, 180^\circ, 270^\circ\}$ for the rotated evaluations.

\subsection{Relative Pose Estimation}\label{subsec:relpose}

\begin{figure}[t]
    \centering
        \includegraphics[width=0.95\linewidth]{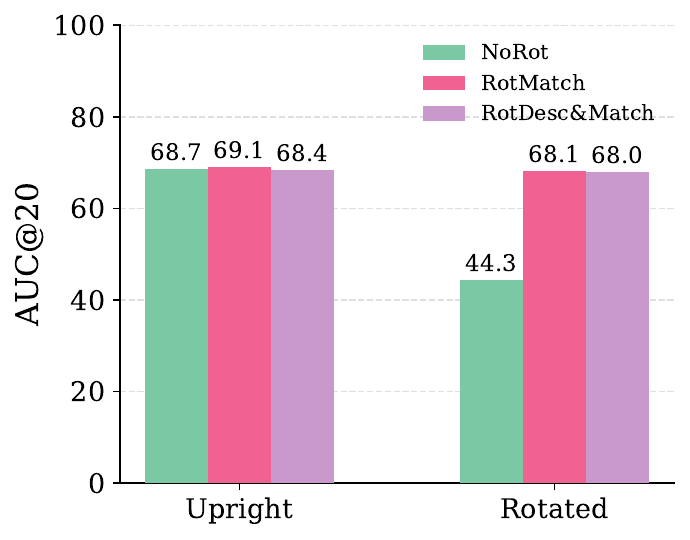}

    \caption{\textbf{ScanNet-1500 performance.} We plot the AUC@20 for our different models.}
    \label{fig:relpose-graph}
\end{figure}

We evaluate relative pose estimation on MegaDepth-1500~\cite{li2018megadepth,sun2021loftr} and ScanNet-1500~\cite{dai2017scannet,sarlin2020superglue}. We use the same evaluation protocol as in prior work~\cite{edstedt2023dkm, edstedt2024roma, edstedt2025romav2harderbetter, loma:2026} and report the performance in \cref{tab:relpose} as well as graphically in \cref{fig:relpose-graph}. We find that incorporating rotation augmentation slightly hurts upright performance. However, it drastically improves performance on rotated image pairs. Interestingly, we find that the drop in performance for the NoRot baseline is much less drastic than that of traditional sparse matchers trained on small datasets. For example, SuperGlue experiences a 52-point drop in AUC@20 on the rotated MegaDepth evaluation, compared to only a 2-point drop for NoRot. In \cref{subsec:ablations} we investigate this further. We find no consistent pattern on the difference in performance between RotMatch and RotDesc\&Match. Empirically, we find that AUC@20 fluctuates by around $\pm 0.1$ due to the stochasticity of RANSAC.
\begin{table}[t]
\centering
\caption{\textbf{Relative pose estimation} on MegaDepth-1500~\cite{li2018megadepth, sun2021loftr} and ScanNet-1500~\cite{dai2017scannet, sarlin2020superglue}. We evaluate both \textit{with} and \textit{without} random rotations.
}
\label{tab:relpose}
\small

\begin{tabular}{l @{\hspace{1em}} rrr @{\hspace{1em}} rrr}
\toprule
Method
& \multicolumn{3}{c}{MegaDepth}
& \multicolumn{3}{c}{ScanNet} \\
\cmidrule(lr){2-4} \cmidrule(lr){5-7} 
AUC$@$ $\rightarrow$ 
& $5^{\circ}$ & $10^{\circ}$ & $20^{\circ}$
& $5^{\circ}$ & $10^{\circ}$ & $20^{\circ}$ \\
\midrule
\multicolumn{7}{@{}l@{}}{\small \textit{Rotated evaluation}} 
\\
SP+SG & 13.3 & 19.1 & 23.8 & 4.6 & 9.6 & 15.0  \\
SP+LG & 13.2 & 19.0 & 23.9 & 3.7 & 7.8 & 12.3  \\
ALIKED+LG & 11.2 & 15.4 & 18.8 & 3.4 & 6.9 & 10.6\\
\midrule
\cellcolor[HTML]{7bc8a4} {NoRot} & 52.4 & 68.8 & 81.2 & 16.8 & 31.2 & 44.3 \\
\cellcolor[HTML]{f06292} {RotMatch} & \bfseries 54.5 & \bfseries 70.9 & \bfseries 82.9 & \bfseries 27.6 & \bfseries 49.6 & \bfseries 68.1 \\
\cellcolor[HTML]{c999cc} {RotDesc\&Match} & 54.2 & 70.6 & 82.8 & 27.4 & 49.5 & 68.0 \\

\midrule
\multicolumn{7}{@{}l@{}}{\small \textit{Non-rotated evaluation}} 
\\
SP+SG & 43.7 & 61.8 & 76.5 & 16.4 & 32.5 & 49.0 \\
SP+LG & 43.8 & 61.8 & 76.4 & 15.9 & 32.1 & 48.9 \\
ALIKED+LG & 48.1 & 65.7 & 79.3 & 14.5 & 28.9 & 43.5 \\

\midrule
\cellcolor[HTML]{7bc8a4} {NoRot} & \bfseries 55.4 & \bfseries 71.7 & \bfseries 83.6 & \bfseries 28.5 & 50.5 & 68.7  \\
\cellcolor[HTML]{f06292} {RotMatch} & 55.3 & 71.5 & 83.4 & 28.4 & \bfseries 50.6 & \bfseries 69.1 \\
\cellcolor[HTML]{c999cc} {RotDesc\&Match} & 55.0 & 71.1 & 83.2 & 28.3 & 50.3 & 68.4 \\

\bottomrule
\end{tabular}
\end{table}

\subsection{Multi-Modal Matching on WxBS}\label{subsec:wxbs}

WxBS~\cite{mishkin2015WXBS} is a challenging matching benchmark consisting of 37 pairs with hand-labeled correspondences. The dataset includes a mix of extreme changes in sensor type, viewpoint, and lighting. We report the mean accuracy (mAA@10px) in \cref{tab:challenging-matching}. Interestingly, our rotation-invariant models outperform NoRot even on the upright evaluation. As WxBS is outside of the training data distribution, this indicates that training with rotation augmentation leads to better generalization, when data is plentiful. We find that traditional sparse matchers fail on the rotated evaluation while NoRot has a less drastic deterioration. 

\begin{table}[t]
\centering
\caption{\textbf{Challenging matching} on WxBS~\cite{mishkin2015WXBS} (mAA@10px) and SatAst~\cite{edstedt2025romav2harderbetter} (AUC@10$^\circ$). SatAst includes arbitrary rotations by default.
}
\label{tab:challenging-matching}
\small

\begin{tabular}{l @{\hspace{1em}} rr @{\hspace{1em}} rr}
\toprule
Method
& \multicolumn{2}{c}{WxBS}
& \multicolumn{1}{c}{SatAst} \\
\cmidrule(lr){2-3} 
\cmidrule(lr){4-4}
& Non-rotated & Rotated & \\

\midrule

SP+SG & 45.6 & 7.3 & 19.8 \\
SP+LG & 46.0 & 10.9 & 12.8 \\
ALIKED+LG & 43.9 & 8.0 & 12.1 \\

\midrule
\cellcolor[HTML]{7bc8a4} {NoRot} & 68.7 & 50.0 & 18.8 \\
\cellcolor[HTML]{f06292} {RotMatch} & \bfseries 71.2 & \bfseries 62.8 & 42.4 \\
\cellcolor[HTML]{c999cc} {RotDesc\&Match} & 65.5 & 62.7 & \bfseries 47.7 \\
\bottomrule
\end{tabular}
\end{table}

\subsection{Astronaut to Satellite Matching on SatAst}\label{subsec:satast}

SatAst~\cite{edstedt2025romav2harderbetter} tests the ability to match images taken by astronauts from the International Space Station with satellite images. The dataset contains 39 corresponding image pairs from EarthMatch~\cite{berton2024earthmatch}, as well as their 90$^\circ$ rotations. We report the AUC@10px for estimated homographies in \cref{tab:challenging-matching}. Matchers trained without rotation augmentation perform poorly due to the large in-plane rotations in SatAst. NoRot only achieves 18.8, comparable with SP+SG, while the rotation-invariant matchers score above 40. 

\subsection{Next Level Matching on HardMatch}\label{subsec:hardmatch}

HardMatch~\cite{loma:2026} is an extremely challenging benchmark that was recently released, featuring 1000 image pairs from 100 diverse categories with hand-labeled correspondences. Although the dataset mostly contains upright images, it also includes pairs with no canonical orientation, such as celestial bodies. We report the mean accuracy (mAA) in \cref{tab:hardmatch}. We find that the models with rotation augmentation outperform the baseline even on the upright benchmark.

\begin{table}[t]
\centering
\caption{\textbf{Taking matching to the next level} on HardMatch~\cite{loma:2026} (mAA@10px).
}
\label{tab:hardmatch}
\small

\begin{tabular}{l @{\hspace{1em}} rr @{\hspace{1em}} rr}
\toprule
Method
& \multicolumn{2}{c}{Rotated}
& \multicolumn{2}{c}{Non-rotated} \\
\cmidrule(lr){2-3} 
\cmidrule(lr){4-5}
mAA@$\rightarrow$
& 5px & 10px & 5px & 10px \\

\midrule

SP+SG & 7.7 & 10.9 & 26.3 & 36.0 \\
SP+LG & 9.8 & 14.3 & 24.8 & 34.8 \\
ALIKED+LG & 6.5 & 9.3 & 26.4 & 35.7\\

\midrule
\cellcolor[HTML]{7bc8a4} {NoRot} & 29.1 & 39.2 & 37.7 & 50.6 \\
\cellcolor[HTML]{f06292} {RotMatch} & 36.0 & 48.6 & \bfseries 37.9 & \bfseries 51.1 \\
\cellcolor[HTML]{c999cc} {RotDesc\&Match} & \bfseries 37.4 & \bfseries 50.6 & 37.4 & 50.8  \\
\bottomrule
\end{tabular}
\end{table}

\begin{figure}[t]
    \centering
        \includegraphics[width=0.95\linewidth]{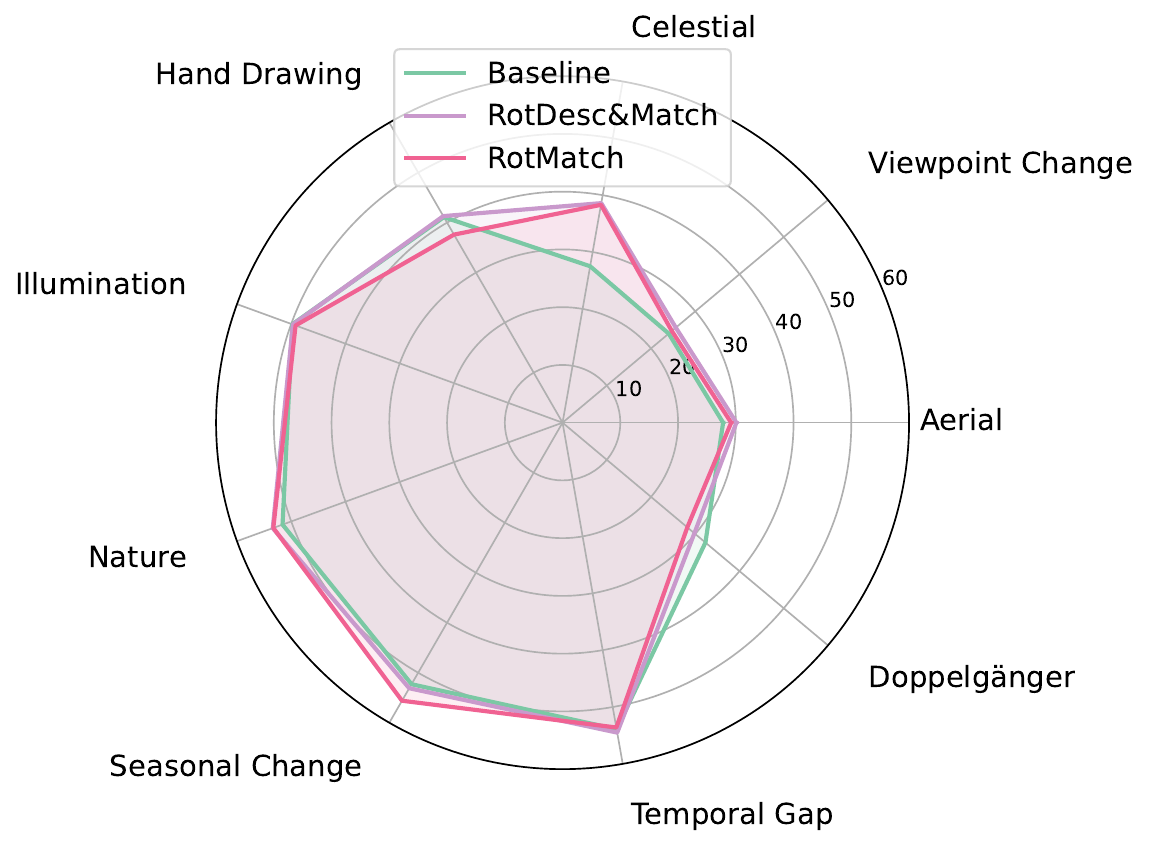}

    \caption{\textbf{HardMatch group breakdown.} We report mAA@10px for different subgroups of the non-rotated version of HardMatch. We find that our rotation-invariant models achieve significant gains on celestial matching.}
    \label{fig:radar}
\end{figure}

\begin{figure*}[t]
    \centering
        \includegraphics[width=0.70\linewidth]{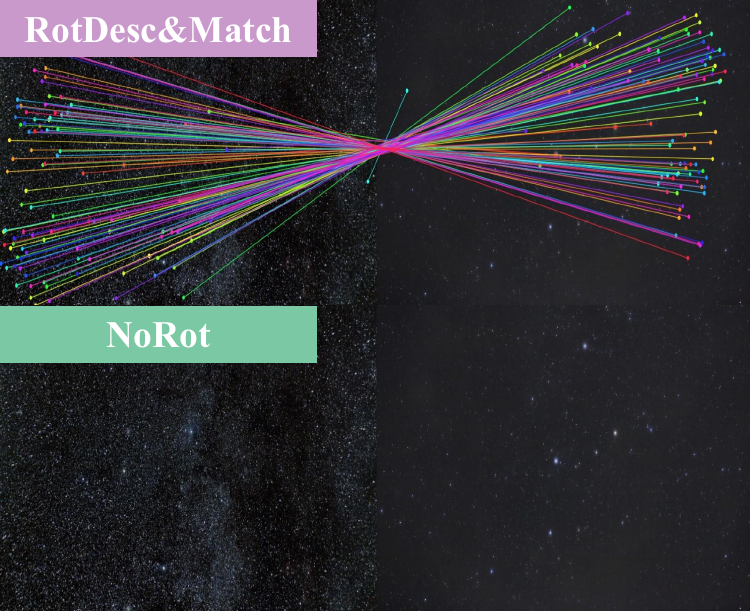}

    \caption{\textbf{Qualitative HardMatch pair.} Rotation invariance significantly improves robustness to image pairs that lack a canonical orientation, such as star constellations. NoRot fails to find any matches above the threshold $\tau$. }
    \label{fig:hardmatch-qual}
\end{figure*}

Motivated by the slight improvement achieved by the rotation-invariant models on the non-rotated HardMatch evaluation, we break down the performance by group in \cref{fig:radar}. We find that performance is largely similar except for a single group, celestial. This group includes images of star constellations and planets displaying natural symmetry to rotations. We illustrate a qualitative example in \cref{fig:hardmatch-qual}.

\subsection{Ablation Studies}\label{subsec:ablations}

\paragraph{Effect of Increased Data.} To investigate the comparative robustness of our baseline compared to other sparse matchers, we study the effect of increased training data. We train a descriptor and matcher pair for 50K steps only on MegaDepth and compare it to one trained on our full dataset. As illustrated in \cref{tab:more-data}, there is a significant difference in robustness to in-plane rotations between the two. Though both are trained without rotation augmentation, the model trained on the full dataset achieves a 52-point increase in AUC@20 on the rotated evaluation set.

\begin{table}[t]
\centering
\caption{\textbf{Effect of increased data.} Relative pose estimation on MegaDepth-1500~\cite{li2018megadepth, sun2021loftr}.}
\label{tab:more-data}
\small

\begin{tabular}{l @{\hspace{1em}} rrr @{\hspace{1em}} rrr}
\toprule
Training Data
& \multicolumn{3}{c}{Non-Rotated}
& \multicolumn{3}{c}{Rotated} \\
\cmidrule(lr){2-4} \cmidrule(lr){5-7} 
AUC$@$ $\rightarrow$ 
& $5^{\circ}$ & $10^{\circ}$ & $20^{\circ}$
& $5^{\circ}$ & $10^{\circ}$ & $20^{\circ}$ \\
\midrule
Only MegaDepth & 53.2 & 69.5 & 82.0 & 16.1 & 22.9 & 29.1 \\
All Data & \bfseries 55.0 & \bfseries 71.3 & \bfseries 83.2 & \bfseries 52.2 & \bfseries 68.7 & \bfseries 81.0 \\
\bottomrule
\end{tabular}
\end{table}

\begin{figure*}[t]
    \centering
        \includegraphics[width=0.70\linewidth]{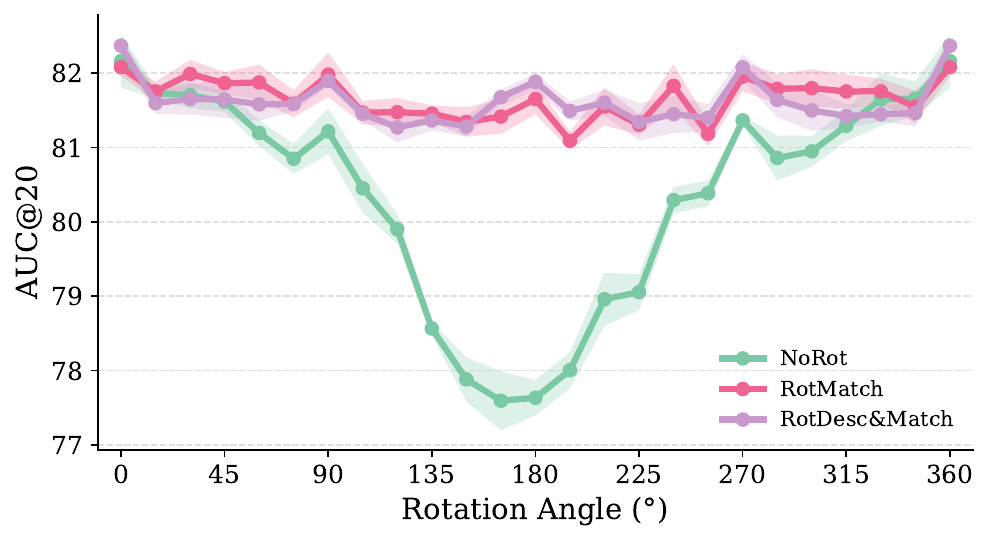}

    \caption{\textbf{Performance for arbitrary rotations.} We report AUC@20 on MegaDepth-1500 at different rotation angles. We average over 5 runs and include a 95\% confidence interval.}
    \label{fig:cont-rotations}
\end{figure*}

\begin{figure*}[t]
    \centering
        \includegraphics[width=0.95\linewidth]{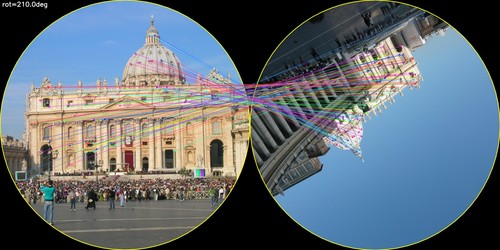}

    \caption{\textbf{Visualizing an arbitrary rotation.} Qualitative matches using circular cropping with  $\alpha^\mathcal{A}=0^\circ$ and $\alpha^\mathcal{B}=210^\circ$.}
    \label{fig:arbitrary-rotations-pair}
\end{figure*}

\paragraph{Arbitrary Rotations.} Here, we study the performance of other rotation angles than multiples of $90^\circ$ which has been used for the other experiments. For a fair comparison, we perform circular cropping for all angles. We keep $I^{\mathcal{A}}$ upright (\ie $\alpha^\mathcal{A}=0^\circ$) and only rotate $I^{\mathcal{B}}$ with a fixed rotation angle $\alpha^{\mathcal B} \in \{0^\circ, 15^\circ, 30^\circ, 45^\circ, \ldots, 360^\circ\}$. We illustrate an example pair in \cref{fig:arbitrary-rotations-pair}. In \cref{fig:cont-rotations}, we illustrate the performance as a function of the rotation angle. We average over 5 runs on MegaDepth-1500. Rotation-invariant models generalize well to other rotation angles than those seen during training with slight jumps in performance for multiples of $90^\circ$. The baseline, NoRot, is robust to small rotations, but struggles as $\alpha^\mathcal{B}$ approaches 180$^\circ$.  
We believe this to be partially due to our data mixture containing aerial data, which naturally provides in-plane rotation, without augmentation.

\begin{figure*}[t]
  \centering
  \begin{subfigure}{0.73\linewidth}
    \centering
    \includegraphics[width=\linewidth]{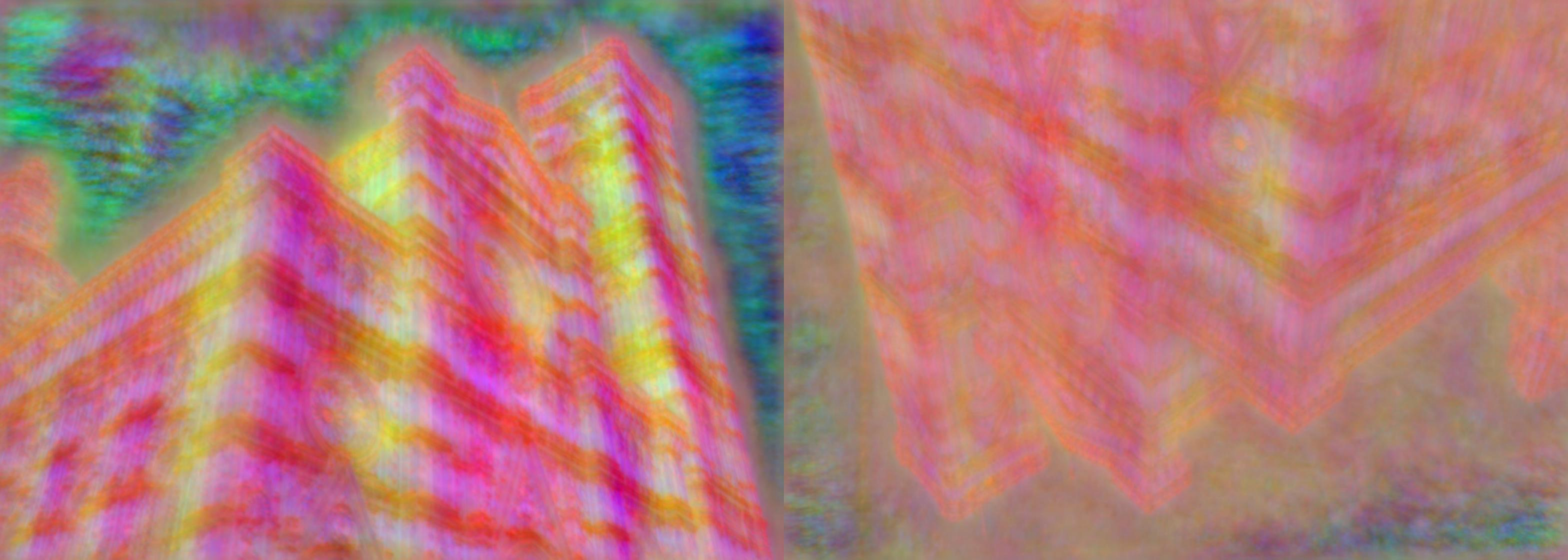}
    \caption{Upright descriptor.}
  \end{subfigure}
  \\
  \vspace{1em}
  \begin{subfigure}{0.73\linewidth}
    \centering
    \includegraphics[width=\linewidth]{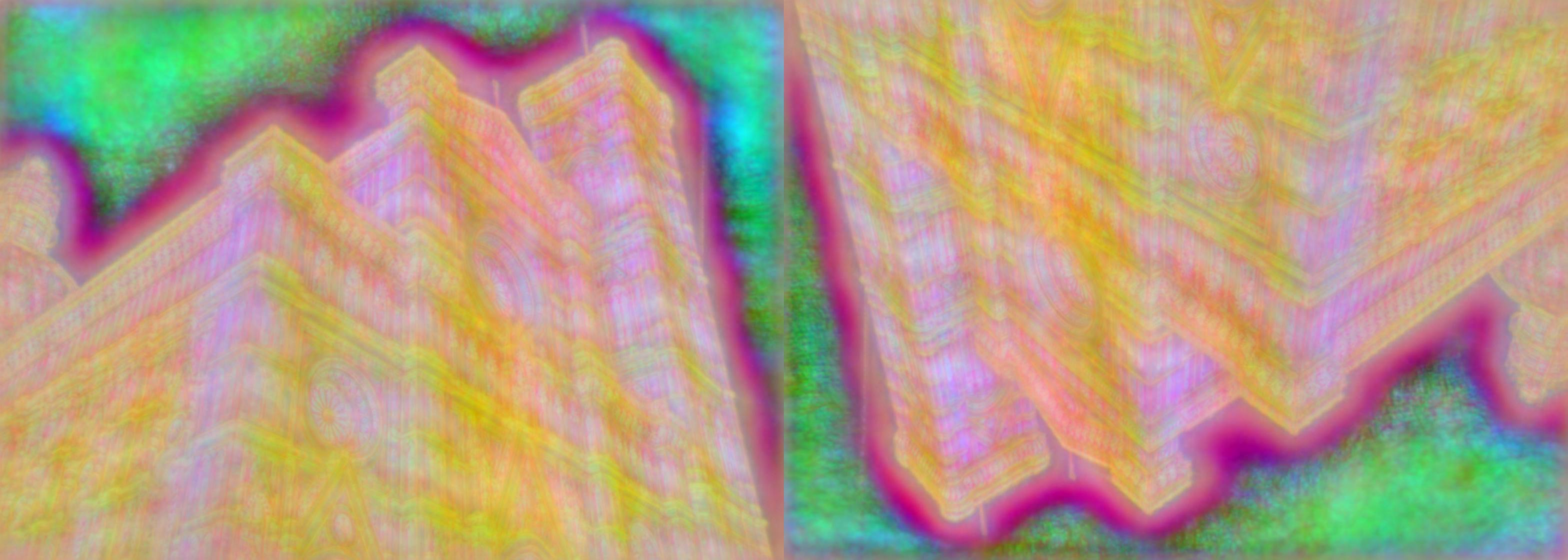}
    \caption{Rotation-invariant descriptor.}
  \end{subfigure}
  \caption{\textbf{Qualitative impact of rotation invariance.} Descriptor features (a) trained with rotation augmentation and (b) without, for an image and its $180^\circ$ rotation.  We visualize the features using SVD, mapping the top three components to RGB.}
  \label{fig:desc-qual}
\end{figure*}

\paragraph{Joint Rotations.} When training the descriptor in RotDesc\&Match we used independent rotation angles $\alpha^\mathcal{A}$ and $\alpha^\mathcal{B}$. We experiment with training a descriptor using the same random rotation on both images, \ie $\alpha^\mathcal{A}=\alpha^\mathcal{B}$ and subsequently training a matcher with independent rotation augmentation (similar to RotMatch). We report the performance on MegaDepth-1500 in \cref{tab:joint-rotations} and compare to our other models. We find the difference to be minimal. A possible slight performance improvement is achieved on rotated evaluations by using joint rotations in the descriptor training. 

\begin{table}[t]
\centering
\caption{\textbf{Effect of joint rotations.} Relative pose estimation on MegaDepth-1500~\cite{li2018megadepth, sun2021loftr}. We compare a descriptor with joint rotations and a matcher with independent rotations with our other models (trained with independent rotations).}
\label{tab:joint-rotations}
\small

\begin{tabular}{l @{\hspace{1em}} rrr @{\hspace{1em}} rrr}
\toprule
Method
& \multicolumn{3}{c}{Non-Rotated}
& \multicolumn{3}{c}{Rotated} \\
\cmidrule(lr){2-4} \cmidrule(lr){5-7} 
AUC$@$ $\rightarrow$ 
& $5^{\circ}$ & $10^{\circ}$ & $20^{\circ}$
& $5^{\circ}$ & $10^{\circ}$ & $20^{\circ}$ \\
\midrule
Joint Rot. Desc.& 54.9 & 71.3 & 83.3 & \bfseries 54.7 & \bfseries 71.1 & \bfseries 83.2 \\
\cellcolor[HTML]{f06292} RotMatch & \bfseries 55.3 & \bfseries 71.5 & \bfseries 83.4 & 54.5 & 70.9 & 82.9 \\
\cellcolor[HTML]{c999cc} RotDesc\&Match & 55.0 & 71.1 & 83.2 & 54.2 & 70.6 & 82.8 \\
\bottomrule
\end{tabular}
\end{table}

\subsection{Qualitative Descriptor Differences}\label{subsec:desc-qual}

We visualize the effect of rotation invariance in the descriptor in \cref{fig:desc-qual}. To allow for meaningful color semantics, we estimate the projection to the top three principal components using SVD in the upright image and use the same projection for the $180^\circ$ rotated version. The descriptor trained with rotation augmentation produces almost identical features for the original and rotated image while the upright descriptor does not.

\section{Limitations}
While providing a comprehensive study to answer our proposed research questions, several limitations persist. 

First, we do not study rotational invariance at the stage of the detector. However, the DaD keypoints that we use are trained with rotation augmentation. 

Second, our networks are not fully rotation-invariant, as can be seen by the reduced performance on rotated benchmarks. One could guarantee invariance by using rotation equivariant network architectures. We view it as interesting future work to pursue this direction, in particular as recent work has found promising designs for utilizing equivariant networks for computational reductions~\cite{bokman2025flopping, nordstrom2025stronger}.

Third, we have not included descriptor training with non-trivial steerers~\cite{bökman2024steerers} which would lead to more distinctive descriptions across rotations. However, our results indicate that having invariance already at the descriptor stage works as well as having invariance only later. It would likely be necessary to modify the sparse matching pipeline to make use of the non-trivial steerer to improve on our results. It is not obvious to us how such a modification of the matcher should be done.
    
Finally, we limit our study to rotations that are a multiple of $90^\circ$. However, we additionally report results for arbitrary rotations in \cref{fig:cont-rotations} to provide further insight into the behavior of the method under arbitrary rotations.

\section{Conclusions}

We studied \textit{who handles orientation} by evaluating the effect of rotation invariance at different stages of a modern sparse matching pipeline. Extensive experiments show that incorporating rotation invariance in the descriptor yields performance comparable to enforcing it only in the matcher. However, we found that rotation invariance is obtained in earlier layers in the former case, allowing faster matching through earlier stopping. We further demonstrate that robustness to in-plane rotations arises naturally from data diversity and that strong performance on arbitrary rotation angles can be achieved even when training on a discrete subset. Interestingly, we found rotation-invariant matchers to outperform their upright counterpart on several non-rotated benchmarks, \eg WxBS. Lastly, we released two rotation-invariant matchers achieving SotA performance on some upright benchmarks and all rotated ones.

\newpage
\section*{Acknowledgements}
This work was supported by the Wallenberg Artificial
Intelligence, Autonomous Systems and Software Program
(WASP), funded by the Knut and Alice Wallenberg Foundation, and by the strategic research environment ELLIIT, funded by the Swedish government. 
The computational resources were provided by the
National Academic Infrastructure for Supercomputing in
Sweden (NAISS) at C3SE, partially funded by the Swedish Research
Council through grant agreement no.~2022-06725, and by
the Berzelius resource, provided by the Knut and Alice Wallenberg Foundation at the National Supercomputer Centre.

{
    \small
    \bibliographystyle{ieeenat_fullname}
    \bibliography{main}
}

\end{document}